# Deep Learned Path Planning via Randomized Reward-Linked-Goals and Potential Space Applications

**Tamir Blum[1]**   **William Jones[1]**   **Kazuya Yoshida[1]**

[1] Department of Aerospace Engineering, Space Robotics Lab, Tohoku University, Sendai 980-8579, Japan
tamir@dc.tohoku.ac.jp, william@dc.tohoku.ac.jp, yoshida@astro.mech.tohoku.ac.jp

## Abstract

Space exploration missions have seen use of increasingly sophisticated robotic systems with ever more autonomy. Deep learning promises to take this even a step further, and has applications for high-level tasks, like path planning, as well as low-level tasks, like motion control, which are critical components for mission efficiency and success. Using deep reinforcement end-to-end learning with randomized reward function parameters during training, we teach a simulated 8 degree-of-freedom quadruped ant-like robot to travel anywhere within a perimeter, conducting path plan and motion control on a single neural network, without any system model or prior knowledge of the terrain or environment. Our approach also allows for user specified waypoints, which could translate well to either fully autonomous or semi-autonomous/tele-operated space applications that encounter delay times. We trained the agent using randomly generated waypoints linked to the reward function and passed waypoint coordinates as inputs to the neural network. Such applications show promise on a variety of space exploration robots, including high speed rovers for fast locomotion and legged cave robots for rough terrain.

## Introduction

Region enabled multi point travel is critical for space exploration robots. Region enabled, referring to being able to travel anywhere within an area, i.e. a lunar crater, and multi point referring to being able to travel to multiple points in succession. Within this crater, it is important to be able to make onboard decisions, such as the best place to look for ice, as opposed to just one point within the crater known beforehand. As the lunar surface has rarely been traversed and the environment poses considerable unknowns, such as surface topology in high precision or the landscape of lunar caves. In such environments, live path planning can continuously update the path as new information is learned and as such increases the likelihood of finding what the agent looking for, like ice. Also, most missions have multiple or regions of interest that each need to be observed rather than just one, such multi point travel is a critical need. This might also apply in other applications like transport and disaster robotics for example.

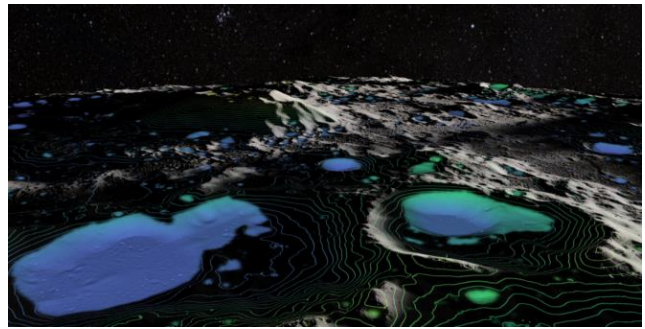

Figure 1: NASA image showing permanently shadowed regions on the lunar surface

The space robotics lab at Tohoku University is developing two novel lunar exploration platforms, a quadruped lunar climbing robot and a high-speed lunar rover that would benefit from such region enabled multi point travel. These platforms require new or more advanced capabilities for space exploration robots, such as faster and more precise onboard obstacle avoidance in order to allow them to travel safely, either within the cave, or at high speeds. As such, deep learning shows potential to enhance capabilities to levels traditional methods cannot match. A first step in this direction is path planning and motion control, as discussed in this paper, and future work will be included later in this paper.

In the design of our path planning neural network policy, we have three main goals. The first goal is to teach a quadruped agent to path plan to the final goal, going between waypoints in the way, autonomously. The agent should be versatile and be able to walk within an area of coordinates within a region, encompassing many different directions and points. The second goal is to have the agent learn to path plan and to walk using only a single neural network, training the two tasks together in end-to-end model free reinforcement learning. The third and final goal is to break the traditional binary reinforcement learning system consisting of only the agent and the environment into a ternary system incorporating the user or "owner" of the agent that can specify waypoints for the robot to travel to.

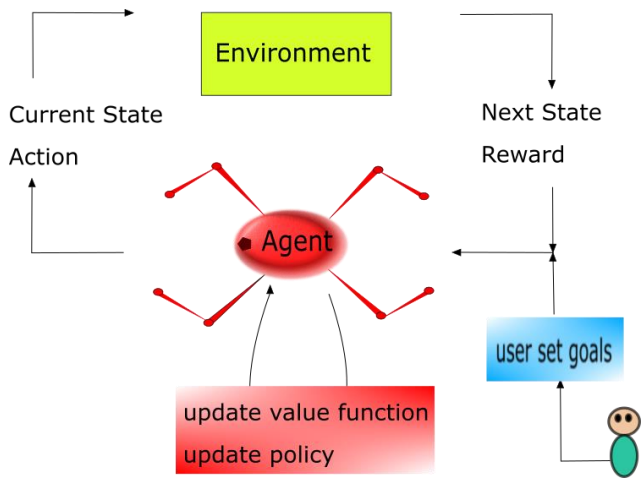

Figure 2: User-Agent-Environment Paradigm for Reinforcement Learning

The ultimate goal of this paper is to take the first step in creating a deep learned path planning and motion control algorithm for space robots. However, the goals mentioned above have a purpose in their own right as well.

The purpose of the first goal is to start to create one of the most basic autonomy requirements of any mobile platform, namely, to allow it to move to designated locations.

The purpose of the second goal is to prove that two fundamentally different, difficult tasks can be handled by a single neural network. This has the advantage of training each functionality simultaneously and can generate complimentary systems. Our results prove promising that additional tasks could also be handled by a single neural network, such as obstacle avoidance, which we will discuss more in further work. More work would have to be done to compare deep learning based approaches to more traditional approaches but recent developments prove it at least warrants investigation.

The purpose of the third goal is to allow for teleoperated command, where a human is added into the mix, yet the robot still commands high autonomy. This is highly useful for space missions which often have scientists determine the points of interest. These points of interests could then be passed to the robot to conduct the mission. Further, in teleoperation cases with high latency, it becomes infeasible or burdensome to manually steer the robot live and so we would like to introduce as much autonomy as possible.

We trained our agent using waypoints randomly generated within a perimeter each episode and linked the reward function always to the current goal waypoint. Additionally, we also made the current and next waypoints observable by passing them to the neural network policy as inputs. The agent controls its joints to "explore" states and through exploring and trying to maximize the reward function, learns to walk and to path plan to waypoints. No data was pre-generated to give to the agent to learn from and no grid or map was generated to aid the path planning.

As a proxy for our own wall climbing robot, which should also be able to walk, we decided to use OpenAI's Gym and OpenAI's Roboschool Ant environment, an 8 degree of freedom robot (Brockman, et al. 2016) (Schulman, et al. 2017) simulated on a flat plane in a 3-D environment and utilizing the Bullet physics engine. This simulated robot is a relatively close simplification of the 12 degree of freedom robot we are developing. Although we trained our policy on a walking robot, we expect the same approach should produce a similar or better result on a wheeled rover considering its dynamically simpler. While we are not currently trying to transfer our learned policies onto the physical system we are developing, we will look to do so in the near future, as will be discussed in future work.

This paper explores and demonstrates the feasibility of real time path planning via a reinforcement learning algorithm that could have applications to lunar and space exploration robots. Further, the simulated quadruped walking agent learns to both walk and path plan at the same time on the same neural network, demonstrating that is possible to teach both high-level and low-level processes together. The agent learns to do this by comparing its current coordinates to the goal coordinates passed in by a user to the system. We explore these with the ultimate goal of empowering a quadruped walking robot to walk and eventually climb on the moon while searching for resources, as well as to empower a high-speed lunar rover to move quickly yet safely. High conservativity is a common problem in space exploration robotics (Otsu, et al. 2018). Perhaps one of the most promising potentials of introducing reinforcement learning into the path planning algorithms for space robotics is that it could allow us to become more liberal in path selection, as the robot can choose to take paths in real time based off sensor information that a more conservative pre-built graph would have avoided.

## Background

### Reinforcement Learning

Reinforced learning refers to a set of algorithms that learn from trial and error. There must be a feedback mechanism to induce learning. In such a case, the agent affects its own observations. This is one of the enticing and powerful features of reinforcement learning as compared to supervised and unsupervised learning. Being able to interact with the environment allows the "agent" to gather its own data and improve itself constantly. Some of the key components of any reinforcement learning algorithm include the policy of how to act, the reward system of instantaneous achievements, the value function of long-term desires, and a model of the world which is optional. There are continuous action spaces, which have a range of acceptable actions (such as a range of acceptable angles for a servo) and discrete action spaces (such as only allowing a limited number of acceptable servo angles). There are also continuous tasks, which are never ending in time, and episodic tasks, which have specified start and end conditions. Recent advancements include running several simulations in parallel and pooling the data collected to speed up learning (Sutton and Barto 2018).

The goal of any policy is to maximize reward, whether a continuous reward or something at the end. Intuitively, maximizing short term gains can often lead to bad results, as many actions that lead to a high instantaneous reward at one time step put the agent in a bad position for the following time steps. Thus, the policy needs to be smart enough to figure out a suitable way to maximize cumulative reward rather than just instantaneous reward. A policy is a function approximator with an input of an array of observations and outputs of an array of certain actions. Reward discounting is a technique to devalue future rewards, in order to factor in that future rewards might be risky and might never be achieved, thus giving more value to near-future rewards that hold less risk. It is important to be careful not to allow positive feedback loops which agents can use to manipulate the reward system and to received excessive unintended rewards while not aiding actual performance. Thus, the reward function has to be chosen carefully and often takes time and many trials to find suitable variables and weights. Exploration vs Exploitation must be balanced in many algorithms. Exploration is the search for new better policies, whereas exploitation is using the current policy to derive maximum benefit. However, in an actual deployed policy, it can be detrimental to include any exploration. As such, some algorithms decrease exploration over time or use an off-policy algorithm. These off-policy algorithms, as opposed to on policy algorithms, have two separate policies, one that is for training and used as a proxy for the real policy meant for deployment. There are also other factors, like entropy which describes the randomness of actions taken, which is often promoted to aid learning and avoid local minima.

**Path Planning**

Path planning algorithms, which tell a robot how to get from point A to point B, are critical in any mobile system. There are numerous ways to accomplish this and different algorithms have different strengths and weaknesses. Some of the most famous path planning algorithms are Breadth First Search, Dijkstra's Algorithm and A* (Dijkstra 1959) (Hart, Nilsson and Raphael 1968) (Moore 1959). Breadth First Search builds a graph from the starting location in all directions. This can often be overly simplistic and inefficient though, causing the algorithm to waste time searching areas fruitlessly. Dijkstra's Algorithm builds on Breadth First Search by introducing costs, thus, some directions are more costly than others and explored proportionally according to cost. This addresses some of the inefficiency issues however does not take into account the agent's actual goal location. A* works similar to the aforementioned algorithms but adds in a heuristic that takes into account the distance to the goal as well as distance from start for any given graph position. It is more efficient as it doesn't bother exploring regions it thinks are more expensive to save resources and time.

Programming the path planning algorithms such as A* can be tedious as the designer has to carefully think about the environment, which might not be well known, in order to introduce information to the system, to determine costs. For many purposes, it is not feasible to have a human to generate a grid for path planning algorithms because it is hard for us to determine the cost accurately and for many others, it is undesirable as it slows down the process and increases development costs among other reasons.

Path planning algorithms are critical part of any autonomous robotic missions that have been investigate for a long time and are a critical step in achieving full robotic autonomy. With recent advancements in artificial intelligence and particularly reinforcement learning, it has become easier to create more customizable algorithms via end to end learning. Using deep learning techniques eliminates the need to add much engineer-chosen information about the environment, and instead all this information is focused on crafting the reward function.

**Related Work**

Much of the early work in the reinforcement learning field was restricted to video games. Recently however, there has been a lot more work with simulated and physical robotic systems. Much of the work still applies to just simulated systems, but slowly more work is coming out on physical systems, usually transferring policies learned in simulation to the physical system. This comes with a challenge of a real to sim "gap" (Tan, et al. 2018). Methods such as include simulating actuator latency help ameliorate these issues.

In regards to prior work done with both motion control and planning, the OpenAI Roboschool Humanoid Flag Runner environment introduces a similar target running task for a humanoid agent to solve based off randomized goals, however, we have not found any papers analyzing the results and methodology, nor any work on non-humanoids, such as quadrupeds, in this regard. As such, it is hard to compare our results, methodology and rationale. Our training algorithms also have significant differences from the above, such as use of a training perimeter, for example.

Other work with planning has been done as well, such as model-free methods using long short term memory (LSTMs), a form of recurrent neural networks, and convolutional neural networks, applied to a variety of applications, mostly 2D in nature (Guez, et al. 2019).

Some work in 3D planning for robots has also been done using hierarchical control to separate the high level planning from the low level motion control, thus containing each in a different neural network (Xue, et al. 2017) (Merel, et al. 2018). This could use the low level controller to pick a referenced walking sequence while the high level controller learns to navigate the robot based off of a known map of the environment (Xue, et al. 2017). It could also use several low level controllers, each with a different reference motion while the high level controller learns to which one to activate (Merel, et al. 2018).

Some other works also reference using reference motions to learn to walk forward in one direction, yet multiple gaits, including later applying those to the real world (Tan, et al. 2018).

Within motion control, there are several possible routes to go. The first is hierarchical. This approach makes use of multiple neural network and divides tasks for each neural network to use (Merel, et al. 2018) (Tan, et al. 2018). Then rather than have one neural network, some strategy is developed to switch between these different neural networks. This could mean having one neural network for walking forward and another for turning or having one for walking and one for trotting. A small number of papers have been written on directly learning on physical systems (Haarnoja, et al. 2018).

A few papers have also focused on symmetric, more animal-like walking (Wenhao, Turk and Karen 2018). Work has been done exploring difficult terrains including obstacles, walls, and forced jumps. This research often involves using a reward function that rewards movement towards the final goal which was fixed during each trial (Heess, et al. 2017).

Many papers on teaching simulated robots to walk focus on training the agent to travel in a single direction, with a single constant reward function (non-varying goal) or use hierarchical control to implement controlled turning (Tan, et al. 2018) (Merel, et al. 2018) (Haarnoja, et al. 2018) (Heess, et al. 2017).

## Training

### Desired Outcome

Our desired agent should understand commands in the form of waypoints and use a learned policy to walk efficiently and quickly to the designated x and y coordinates (within a specified threshold). It should determine walking characteristics, such as which direction, how far it needs to go and how fast it should go, by itself. It should be able to walk anywhere within an area, not just to just one point or in just one direction. It should also be able to walk to multiple points in succession.

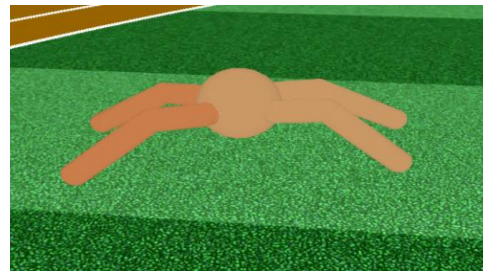

Figure 3: Roboschool Ant agent's body.

### Simulation Environment

The focus of this paper does not demand an extremely complicated environment. Having one could lead to confounding issues. As such, we use a flat plane without any obstacles in a 3D world with an 8 degree of freedom ant-like robot. See future work for possible follow up directions with more complicated environments.

### Agent

We trained on a four-legged ant like agent created by OpenAI in their Roboschool platform. We left the agent XML, which defines the robot in terms of body and joint properties, unchanged, however changed other properties such as the "observations" made (i.e. the information passed into the neural network). This agent has 8 joints, yielding an 8-element action array as the output of the neural network policy used to control each joint. Each of the joints is a hinge type joint with only one degree of freedom. The joints are controlled continuously but have some constraints, such as range of motion. This agent serves as a simplified proxy for our lab's wall climbing robot in development, a 4 legged walking robot being created for cave exploration and resource prospecting on the moon.

### Policies

We created 5 policies to validate our training methodology. We tested various permutations of our training methods, including our waypoint randomization training strategy as well as whether to add additional information to the state array, such as goal information, random noise or nothing. The policies are defined below:

Training Styles:
- Single Point Training: Single point policies were trained to go to the center of the training perimeter, the coordinate (10,10) each time
- Waypoint Training: Waypoint training policies were trained with 4 waypoints randomly generated within the training perimeter at the start of each episode

Information Styles:
- State Info: The inputs to the neural network consisted of a 25 element array consisting of state information
- State Info + Noise: The inputs to the neural network consisted of a 29 element array, with 25 elements containing state information and 4 elements containing randomized noise generated at the start of each episode
- State + Goal Info: The inputs to the neural network consisted of a 29 element array, with 25 elements containing state information and 4 elements containing goal coordinates for two waypoints (current and next)

Policy 1: Single Point Training x State Info
Policy 2: Waypoint Training x State Info
Policy 3: Waypoint Training x State Info + Noise
Policy 4: Single Point Training x State + Goal Info
Policy 5: Waypoint Training x State + Goal info

### Training Algorithm and Technique
We train using an algorithm that generates four waypoints randomly at the start of each episodes. If the robot fails to get to any waypoint, the episode is terminated after 10 seconds and a new one is begun. If the robot is able to successfully reach a waypoint, the robot has until 20 seconds to get to the next waypoint and so on for the four waypoints.

In order to give the robot a head start in learning how to walk, we generated a training perimeter in the (+X, +Y) quadrant for which all the waypoints to be generated within. We believe this helped to slightly decouple the problem of learning to walk and learning to path plan, as it can focus solely on walking towards the (+X, +Y) quadrant during the early stages without having to orient itself. We chose to do this after observing that the quadruped agent is not the most agile, and can have some difficulties turning, particularly during early stages of learning to walk. This has also been found by other studies (Heess, et al. 2017). See future work for possible modifications to the agent's body. To ensure this head start, we chose to use a 25m$^2$ square region. We centered this training perimeter around the coordinates (10,10). Four coordinates from within this training perimeter were randomly chosen at the start of each episode.

### Reward and Reinforcement Learning Algorithm
In order to encourage the agent to accomplish the desired outcome, we had to create a reward function that was dependent on the goal. The parameters and weights were consistent amongst all the training episodes, however, the goal which change with each episode and even within the episode if the agent was able to reach a waypoint.

(1) $R(G) = V_G - P$

We penalized the agent modestly for energy usage. The training was done using OpenAI's BASELINES ACKTR reinforcement learning implementation. Many of the penalty terms were taken from the default OpenAI Roboschool Ant platform but rescaled to prioritize velocity more (Brockman, et al. 2016). ACKTR stands for Actor Critic using Kronecker-Factored Trust Region (Wu, et al. 2017). We chose ACKTR as it seemed to perform relatively well on simulated quadruped robots in previous studies (Fujimoto, van Hoof and Meger 2018).

## Experiments
### Setup
We ran a series of experiments in order to validate the necessity of randomized reward-linked goals during training and the presence of the goal information in the observation array. We also ran some experiments to show the necessity of using deep learning by comparing performance of a shallow neural network to a deeper one.

### Points of Interest
We ran experiments to find the effects of:
1) Observable Goals
2) Inclusion of Observable Noise Elements
3) Randomized Waypoint Training
4) Single Point Training
5) Both Observable Goals and Randomized Waypoint Training
6) Shallow vs Deep Neural Network

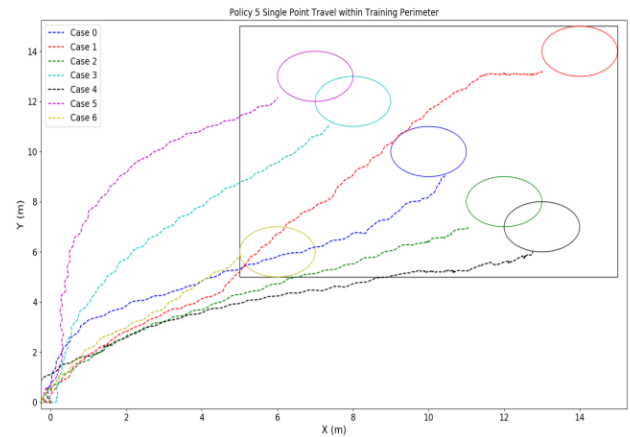

Figure 4: Successful trials using policy 5 to travel to different test cases. Episode terminated once the agent entered a circle indicating the boundary.

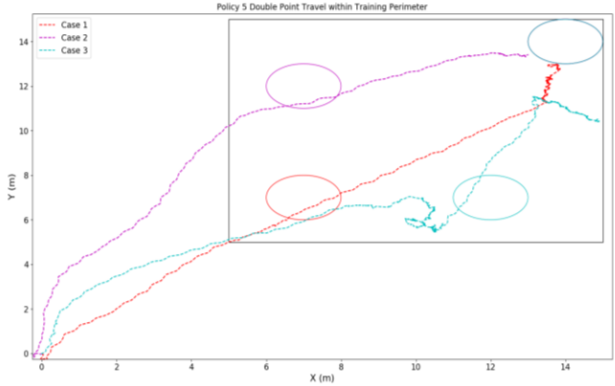

Figure 5: Two-point test cases showcasing some of the policies strengths and weaknesses.

## Neural Network

### Network Parameters

One of the difficult parts of reinforcement learning is the selection and tuning of network parameters. Often, these parameters must be selected heuristically, via trial and error, or via general rule of thumbs that various people. We decided to use the default learning parameters in OpenAI Baselines that are pre-tuned for walking in one direction.

The neural network consists of six hidden layers with 128 hidden units per layer. The input array has 29 elements while the output has 8 elements.

### Array Information

In our case, we chose to use a minimalist state array with 25 elements for the input to the neural network. This consisted of 8 joint angular positions and 8 joint angular speeds; the x, y and z position of the body; the vectorized velocity $V_x$, $V_y$ and $V_z$ of the body; and the roll, pitch and yaw orientation of the body.

The goal array consists of two waypoints, the current waypoint and the next waypoint. It is thus a 4 element array, where each waypoint consists of an x position and y position element.

### Training Waypoint Randomization Algorithm

In order to achieve path planning, we introduced observable goals by passing the information to the neural network along with implementing a randomized goal-linked reward function algorithm for training.

### Algorithm Terms

TP = training perimeter; M = number of waypoints; Alg = learning algorithm; BX, BY = boundary threshold for X and Y; $t_{inc}$ = episode length increase; $t_{ep}$ = initial episode length; t = current time; e = episode counter; GX, GY = goal X and Y coordinates; PX, PY = current X and Y coordinates of robot body; R(G) = reward function w.r.t G; G = goal array; env = environment

**Algorithm 1**

**for** e $\in$ {1, ..., N} **do**
   Generate M randomized waypoints within TP
   Set G
   Set R(G)
   Run simulation with Alg, env
   **if** abs(PX-GX) < BX **and** abs(PY-GY) < BY **then**
     Update R(G)
     Update G
     $t_{ep}$ += $t_{inc}$
   **if** t > EL **then**
     End episode

## Results

The policy trained via the randomized reward-linked waypoint algorithm, in conjunction with observable goals passed into the neural network input, policy 5, was able to learn path planning and motion control to a degree that the other policies, either trained with just one or none of these features, could not. Table 1 showcases the large gap between policy 5 and the rest of the policies. We tested each policy for 10 trials on a two-point path, first going to the waypoint (7,12) before going to the final goal of (14,14). Policy 5 was able to achieve a 90% success ratio while the other policies could not pass once. We also tested policy 5 on six single point test cases with 10 trials per test case. The policy was able to achieve high success ratios on most of the points.

The data supports the feasibility of region enabled multi point travel through deep reinforcement learning. It also shows that multiple tasks, i.e. path planning and motion control, can be done on a single neural network and that it is possible to put a user into the system to give waypoint and goal commands to the agent.

However, policy 5 had some weaknesses that should be addressed in future work. This is evident from the non-perfect success ratio in the tables and in Figure 5. The algorithm should be improved until a perfect success ratio can be achieved for within the region area. We will hypothesis as to possible issues and potential fixes in the future work area.

Table 1: Test results for all policies against a two-point path test

| Policy | 1 | 2 | 3 | 4 | 5 |
|---|---|---|---|---|---|
| Success Rate | 0 | 0% | 0% | 0% | 90% |

Table 2: Success ratio for single-point path test cases (Figure 5)

| Test Case | Goal Coordinate | Success Ratio |
|---|---|---|
| 0 | (10,10) | 100% |
| 1 | (14,14) | 70% |
| 2 | (12,8) | 70% |
| 3 | (8,12) | 80% |
| 4 | (13,7) | 30% |
| 5 | (7,13) | 50% |
| 6 | (6,6) | 80% |

Table 3: Success ratio for two-point path test cases (Figure 4)

| Test Case | Waypoint → Final Goal | Success Ratio |
|---|---|---|
| 1 | (7,7) → (14,14) | 80% |
| 2 | (7,12) → (14,14) | 90% |
| 3 | (12,7) → (14,14) | 0% |

**Policy Analysis**

Policy 1 established a baseline to compare the other policies to. It was trained to walk to the same point every time and could not walk anywhere else.

Policy 2 demonstrated that waypoint training on its own cannot teach the agent to path plan. Such policies would typically walk towards the center of the training perimeter and then explore the area hoping to randomly find a waypoint.

Policy 3 could be useful for transfer learning in future work to compare learning both motion control and path planning at the same time, vs learning to walk first then learning to path plan after. Adding the noise did not see any increase in performance.

Policy 4 showed that it is not sufficient to give the agent just observable goals without also using the randomized waypoint training. It also showed interesting behavior on not walking to walk once the goal information changed. When the goal information was to walk to waypoints at (10,10) it would walk fluidly, but when either one or both of the waypoint goal info coordinates changed, it would not move.

Policy 5 was the only one that learned to path plan as well as motion control. It shows the necessity of both observable goals and randomized waypoint training.

Preliminary results in training networks of different sizes showed shallower networks having difficulties traveling to multiple points, possibly due to the inability to store enough information in the smaller neural network, but this should be studied further.

**Future Work**

Much more still needs to be done before we can use deep learning-based path planning and motion control on real systems for space exploration. Based off the results shown in this paper, we believe that further work in this area is warranted and promising in several areas.

**Algorithm Future Work**

Several upgrades could be made to ameliorate the weaknesses of the training algorithm that could address some paths being harder than others and non-perfect success ratios. One possible cause is that the training isn't standardized enough. This could result from the lack of guarantee on the distance between waypoints. One example is to ensure that during training, each of the generated waypoints is at least a certain distance from the others. Another possible cause might be the entropy being too high in the final policy. Further, giving the agent tasks that it failed more often during training, such as is done in the Roboschool Flag Runner environment could also potentially help. Newer algorithms seem to outperform ACKTR, such as TD3 (Fujimoto, van Hoof and Meger 2018). Use of these could also help. Further, full randomization of waypoints might not be needed during training. Perhaps a small number of training cases, iterated over, would be sufficient if they are representative enough of all the possible (desired) paths. More work could also be done on different training perimeters, i.e. it could be enlarged, changed into various shapes or made global/completely removed. Perhaps more complicated neural network architectures, such as those utilizing LSTMS or CNNs, could help in addressing these issues (Guez, et al. 2019). See future work.

**Tangential Future Work**

Future work for deep learned path planning and motion control contained in a single neural network includes exploring more environments similar to a lunar environment. This could mean including obstacles and varying terrain, such as including hills/craters and small rocks. Some of these obstacles could be surmountable, like the small rocks, while others might require going around them. Work can also be done in incorporating live obstacle detection into the neural network, to feed data necessary for path planning in complex environments.

These changes would allow the robot to learn much more complex path planning techniques, considering much more factors like time and danger. Work on a more complicated robot, such as one closer to our robot in development, could yield more interesting walking gaits and would be the first step in trying to develop algorithms that could be transferred over to a physical system.